# Opportunities for Persian Digital Humanities Research with Artificial Intelligence Language Models; Case Study: Forough Farrokhzad

Arash Rasti Meymandi^, Zahra Hosseini^, Sina Davari, Abolfazl Moshiri, Shabnam Rahimi-Golkhandan, Khashayar Namdar, Nikta Feizi, Mohamad Tavakoli-Targhi*, Farzad Khalvati*

University of Toronto, ON, Canada

^Co-First Authors

*Co-Senior Authors

## Abstract

This study explores the integration of advanced Natural Language Processing (NLP) and Artificial Intelligence (AI) techniques to analyze and interpret Persian literature, focusing on the poetry of Forough Farrokhzad. Utilizing computational methods, we aim to unveil thematic, stylistic, and linguistic patterns in Persian poetry. Specifically, the study employs AI models including transformer-based language models for clustering of the poems in an unsupervised framework. This research underscores the potential of AI in enhancing our understanding of Persian literary heritage, with Forough Farrokhzad's work providing a comprehensive case study. This approach not only contributes to the field of Persian Digital Humanities but also sets a precedent for future research in Persian literary studies using computational techniques.

## 1. Introduction

### 1.1. Digital Humanities: Importance and scope

Individuals have processed and documented human experience through philosophy, literature, religion, art, music, history, and language. The humanities can be defined as the study of how humans understand and record their world. Digital humanities are regarded as the intersection of computing and the discipline of humanities, where digital technologies are used to analyze and model the records of human experience [1].

Digital humanities leverage a wide range of digital methods to gain an overview of humanities documents and records. Traditional approaches such as text markup, topic modeling, network analysis, discourse analysis, virtual modeling, simulation, and aggregation of materials [2] are improved by Artificial Intelligence (AI). One major goal of AI is to train intelligent agents capable of performing ordinary human tasks. A subfield of AI focusing on algorithms specialized in reading and understating text corpus is called Natural Language Processing (NLP). Two interwoven subfields of NLP are known as Natural Language Understanding (NLU) and Natural Language Generation (NLG). NLU implements various techniques for understanding the meaning of the text generated by humans while NLG generates text that is understandable by humans [3]. In general, NLP techniques have allowed for objective, meticulous, and large-scale analysis of the written language.

## 1.2. Exploring Persian Poetry with AI and NLP: The Case Study of Forough Farrokhzad

Building upon our research objective, we specifically chose Forough Farrokhzad as our case study due to her pivotal role in modern Persian literature and her profound impact on the evolution of Persian poetry. Farrokhzad's work, renowned for its bold themes, emotive depth, and distinctive style, presents an ideal corpus for applying and testing the capabilities of NLP and AI models. Her poetry, characterized by a unique blend of personal experiences, cultural nuances, and linguistic richness, offers fertile ground for exploring the intersections of language, emotion, and cultural context [4]. By focusing on Farrokhzad's oeuvre, this study not only pays homage to her literary legacy but also leverages her work as a microcosm to gain deeper insights into the broader spectrum of Persian poetry. This case study approach aims to provide a comprehensive understanding of the stylistic and thematic elements in Farrokhzad's poetry, setting a precedent for future research in Persian literary studies using advanced computational techniques.

## 2. Literature Review: NLP Methods and Their Implications for Digital Humanities with a Focus on Persian Language

This section delves into the various facets of NLP, starting with supervised learning approaches. Supervised NLP, a cornerstone of modern computational linguistics, plays a pivotal role in deciphering complex language structures, especially in rich literary traditions including Persian poetry. We begin by examining the key methodologies, significant studies, and diverse applications of supervised NLP, providing insights into how these techniques are shaping the analysis and interpretation of Persian literary texts. Following this, we will transition to unsupervised learning methods, which harness the power of algorithmic analysis without the need for labeled datasets, uncovering latent themes and stylistic features inherent in Persian texts. Table 1 showcases the application instances of various supervised and unsupervised NLP methods in digital humanities.

*Table 1- Interplay of supervised and unsupervised NLP Methods and their applications in digital humanities*

| NLP Learning Paradigms | | | |
|---|---|---|---|
| Supervised Learning | | Unsupervised Learning | |
| Methods | Application instances | Methods | Application instances |
| • Named entity recognition | o Metaphor detection | • Clustering | o Key passages detection |
| • Part of speech tagging | o Lexicon-based emotion analysis | • Semantic change detection | o Modeling literary styles |
| • Sentiment analysis | o Hate speech detection | • Topic modeling | o Topic-related passages analysis |
| • Semantic role labeling | o NER in historical research | | o Analyzing evolution of social discussions |
| • Semantically related words | | | |
| • Word sense disambiguation | | | |

## 2.1. Application of Supervised NLP in the Persian Language

The analysis and interpretation of intricate literary works, particularly in poetry, are significantly aided by supervised NLP methodologies within the domain of Digital Humanities. This section delves into the application of such methods to Persian language, highlighting their multifaceted contributions.

Supervised learning hinges on "labeled" data, where the model extracts knowledge from annotated examples. This empowers tasks like Named Entity Recognition or NER [5], [6], which pinpoints crucial entities such as names and places. For instance, in a Hafez verse, " حافظ در شیراز به دنیا آمد," NER identifies 'حافظ' and 'شیراز' as significant entities, paving the way for in-depth semantic analysis. Another vital supervised technique, Part of Speech (PoS) tagging, assigns grammatical categories to each word, facilitating syntactic parsing and aiding the overall comprehension of sentence structures in Persian literary analysis [7].

Sentiment analysis has received considerable attention, addressing the classification of emotions within Persian texts. However, the scarcity of annotated data remains a challenge [8]. This method delves deeper, gauging the emotional undercurrent of literary pieces, particularly insightful in the nuanced realm of poetry [9], [10]. Furthermore, Semantic Role Labeling (SRL) clarifies roles within sentences (e.g., "مریم سیب خورد"), while Word Sense Disambiguation (WSD) interprets intended meanings of words with multiple senses. These methods, along with Lexical Substitution, are indispensable for comprehending the layered meanings within Persian texts [11].

The practical applications of these methodologies are manifold, extending to metaphor detection, which unravels the figurative language often prevalent in Persian poetry [12]. Lexicon-based emotion analysis and hate speech detection utilize supervised NLP to examine the sentiment and sociolinguistic implications of texts, offering insights into the cultural and historical contexts of the literature [13]. Lastly, the prediction of readers' preferences by analyzing sentiment arcs exemplifies the capacity of supervised NLP to marry linguistic analysis with reader-response theory, highlighting the interplay between textual composition and audience reception [14], [15].

While English resources dominate NLP, Persian language research is expanding. For example, an adaptable framework has been developed for converting Persian prose into different formats of ancient Persian poetry, utilizing parallel prose-to-poem samples and datasets such as Ganjoor [16]. This pretrained model, initially using unsupervised methods, later incorporates supervised learning for precise translations from prose to poetic forms. Similarly, supervised methods have been deployed for poet identification using the rich Ganjoor dataset, leading to promising performance in authorship attribution tasks [17]. In exploring thematic similarities within poetic verses, supervised approaches leverage pretrained multilingual BERT models to analyze multiple-choice questions based on poetic content, with notable accuracy [18], [19].

Another study tackles the challenges of Persian news text classification by introducing an innovative BiGRUACaps method, combining Bidirectional Gated Recurrent Units with an attention mechanism and Capsule Networks. This method outperforms traditional Long short-term memory (LSTM) models, yielding an F1 score of 0.8608, further advancing Persian NLP tasks [20]. Moreover, [21] explores the realm of metrical pattern identification in poetry. The study reveals that transformer-based models, initially designed for semantic tasks, can effectively retain structural information crucial for metrical analysis in poetry, performing well in multilingual settings and suggesting potential for cross-lingual transfer that could be beneficial in analyzing Persian poetry with its unique linguistic structures and poetic forms.

In addition to fully supervised learning for NLP tasks in Persian text, semi-supervised and weakly supervised learning methods have also been used in the literature. While both offer possibilities for handling limited labeled data, they differ in their approaches. Weakly supervised learning utilizes noisy or incomplete labels, as exemplified by sentiment analysis on colloquial Persian using adversarial training [22] and significant news detection with a pioneering dataset [23]. In contrast, semi-supervised learning leverages a small amount of high-quality labeled data combined with a much larger volume of unlabeled data, as demonstrated by the Persian NER model developed by [24]. Understanding these distinctions allows researchers to make informed choices based on specific data availability and task requirements when working with NLP in resource-constrained languages like Persian.

In conclusion, supervised NLP has become instrumental in unlocking the complexities of Persian language within Digital Humanities. From named entity recognition and sentiment analysis to metaphor detection and authorship attribution, its applications continue to evolve, enriching our understanding and appreciation of Persian literature. Continued development of annotated datasets and exploration of novel techniques promise even more exciting advancements in the years to come.

## 2.2. Unsupervised NLP Methods in Persian Language

Building upon the foundation of supervised NLP methods, the domain of unsupervised NLP offers an expansive toolkit for analyzing Persian literature without the need for labeled data. Clustering algorithms, for example, can group texts from various historical periods into dynastic clusters, allowing for an organized review of literary progressions. Semantic change detection addresses the evolution of word meanings over time, as illustrated by the shifting use of the term "خداوند" from ancient royal connotations to its modern interpretation [25], [26].

Topic modeling stands out as a particularly powerful unsupervised NLP method, utilizing algorithms such as Latent Dirichlet Allocation (LDA) [27] and Non-Negative Matrix Factorization (NMF) to distill large corpora into discernible themes. This technique aids in summarizing, categorizing, and visualizing content, enriching our understanding of literary works [28], [29]. Applications of these unsupervised methods are varied and insightful. They range from identifying pivotal passages that enhance comprehension of literary pieces to modeling the narrative styles of different authors and analyzing topic-related phrase patterns. Moreover, they play a crucial role in tracking the evolution of social discussions, as observed in discourse shifts during events like the COVID-19 pandemic [30], [31].

Incorporating unsupervised NLP into Digital Humanities facilitates nuanced literary analysis, such as the identification of key passages within a body of work [32]. This process involves detecting sections that are particularly informative or evocative, enhancing our comprehension of the text. Another sophisticated application is the modeling of literary styles, which analyzes the narrative techniques of different authors, capturing their unique voices and storytelling methods [33]. Additionally, topic-related passage analysis employs unsupervised NLP to delve into the text corpora, uncovering phrase patterns that align with specific themes, thereby offering a thematic dissection of the content.

[34] presents a quantitative analysis contrasting the frequency of nation-related references within the sentences of various documents to the overall length of these documents, as measured in sentences. The visualization underscores a non-linear relationship, where some shorter texts may feature a higher frequency of nation-related content than longer ones, highlighting the varying thematic focus on the concept of the nation across the corpus. This discrepancy points to the necessity of considering both the quantity and the density of thematic references for a comprehensive understanding of a document's thematic preoccupations [34].

Unsupervised NLP has been pivotal for projects such as ParsBERT, which comprehends the Persian language through extensive training on diverse datasets. ParsBERT's adaptability is demonstrated in its application to various NLP tasks, including sentiment analysis and text classification [35]. The generation of fake Persian poems also represents a creative use of unsupervised learning, where models are trained on classic Persian poetry to produce new compositions that blend different cultural and stylistic elements [36].

These advanced unsupervised NLP techniques are instrumental in expanding the analytical capabilities within the realm of Persian literature, allowing scholars to dissect texts in a manner that reveals underlying structures, styles, and sentiments that might not be immediately apparent.

# 3. Methodology

## 3.1. Dataset

Description of Forough's poetry and the five available books as the primary materials for conducting topic modeling.

1. *Asīr* (*Captive*). 2nd ed. Tehran: Amīr Kabīr, 1334/1955.
2. *Tavalludī dīgar* (*Another Birth*). 4th ed., Tehran: Murvārīd, 1348/1969.

3. *ʿIṣyān* (*Rebellion*). 4th ed., Tehran: Amīr Kabīr, 1348/1969.
4. *Dīvār* (*The Wall*). 5th ed., Tehran: Amīr Kabīr, 1352/1973.
5. *Īmān biyāvarīm be āghāz-e faṣl-e sard* (*Let Us Believe in the Beginning of the Cold Season*). 7th ed., Tehran: Murvārīd, 1368/1989.

## 3.2. Preprocessing

To prepare the data, we first compiled a list of common Persian stop words, working closely with humanities experts to ensure its relevance to our dataset. Table 2 illustrates the list of stop words removed from the dataset:

*Table 2- List of the removed stop words*

| | | | | | | | | | | | |
|---|---|---|---|---|---|---|---|---|---|---|---|
| ما | دگر | نیست | و | آن | مرا | میکند | کش | همه | به | او | میکنی | نیست |
| گر | دیگر | کس | داشت | این | چگونه | با | تو | است | رسان | برای | شده | کشید |
| اگر | تا | میکردم | دار | اما | آور | ده | یا | کرد | رساند | باز | میکنند | چقدر |
| هر | گرفت | میکردیم | چون | همچو | آورد | داد | میکنم | کن | زیر | مثل | میتواند | بر |
| را | گیر | من | کیست | همچون | میشود | این | هست | بود | چرا | شاید | زد | که |
| ز | باید | از | چیست | خود | میان | در | است | باش | هم | آیا | زدن | میرفت |

Next, we used the Natural Language Toolkit (NLTK) library [37] to clean the data by removing punctuation marks and redundant white spaces. We used Hazm [38] and PersianStemmer [39] libraires, which are specifically designed for handling Persian language text, for stemming and lemmatization. Finally, we applied the NLTK library's tokenizer to break down the text into smaller units.

**Stemming**: Stemming is a process in NLP where words are reduced to their root or base form by removing suffixes. For example, "رفته‌ام" and "رفته" would both be stemmed to "دو." Stemming simplifies words to their core, but it may result in some words that are not real words. For instance, "می‌خوانمش" would be stemmed to "خوانم" or just "خوان".

**Lemmatization**: Lemmatization, on the other hand, is a more sophisticated method that reduces words to their base or dictionary form, known as lemma. Unlike stemming, lemmatization ensures that the resulting words are actual words. For example, "دویدن" and "دونده" would both be lemmatized to "دو" and "می‌خوانم" would be lemmatized to "خواندن".

We noticed that the stemming and lemmatization steps struggled with more complicated words such as "می خوانمش", treating such words as unique tokens.

## 3.3. Bag of Words Analysis

## 3.3.1 Frequency Analysis

By using stemming and lemmatization, we standardized the words in the text, making it easier to analyze and identify patterns. Then, we counted the frequency of each word to create simple histograms, providing insights into the most used words and recurring themes in Forough Farrokhzad's poetry. Histograms and word clouds of each poem were then plotted.

### 3.3.2 Clustering

Clustering helps us organize the poems based on how similar they are, revealing common themes or styles. It is similar to grouping poems that share certain word patterns. For instance, poems in one cluster might convey similar concepts or use similar words. Thus, clustering makes it simpler to understand the different flavors and styles in Forough Farrokhzad's poetry. In this work, we cluster poems in two ways: first, by the top five most frequent words in each poem, revealing dominant themes. Second, based on n-grams, capturing nuanced relationships between words for richer insights. Once the word embedding was completed using each method, K-Means clustering algorithm [27] was applied to group similar poems into the same clusters. We chose the number of clusters to be 4 based on experiments since it illustrated more meaningful clusters for this dataset. For the visualization, we used principal component analysis (PCA), plotting the clustering result using the first two principal components for each poem.

### 3.3.2.1 Frequency-Based Clustering (Top five most common words)

In the Frequency Clustering analysis, we group the poems based on the frequency of their top five most common words. The choice of five ensures a balance between capturing significant words and maintaining specificity. This was done similar to Bag of Words (BoW) text embedding method where 1442 single words in the corpus were used. The only difference with typical BoW method is that we only considered the top five most common words in each poem, assigning 0 to the rest of the words in the embedding.

### 3.3.2.2 N-Gram Frequency-Based Clustering and Similarity Analysis

Frequency-based analysis in the previous section was based on individual words, which does not consider the sequence of words. N-Grams BoW embedding method considers groups of words that appear together in the text and thus, predicts the probability of a word given n-1 previous words. This considers words in relation to words before and after in the text, adding context to the analysis. N-Grams can be single words (unigrams) (similar to basic BoW method), pairs of words (bigrams), or sets of three words (trigrams) or more found in the text. We chose trigrams (n=3) to capture the relationship between a word and immediate words around it, which gives a more complete picture of the word's meaning in the context. We only regarded sets of three words as a trigram if they appeared in a verse (line) of a poem.

For example, a sample trigram would be:

A line of poem: بر لبانم سایهای از پرسشی مرموز

A sample Trigram: [('سایه', 'پرسش', 'مرموز')]

In total, 2907 sets of unique three words were captured in the corpus. Next, we calculated the histogram of each poem using the trigrams based on how often they appear in the poems. Similar to unigram analysis, clustering was done using K-means algorithm (4 clusters) followed by PCA analysis for visualization.

Using the trigrams of the poems, we also looked at how similar the poems are to better understand which poems share common phrases or themes. We used Cosine Similarity metric, which measures the cosine of the angle between two vectors (word embeddings) in a high-dimensional space. Equation 1 is the formula for Cosine Similarity between two vectors A and B:

$$Cosine\ Similarity(A, B) = \frac{A.B}{||A|| \times ||B||} \quad \text{Equation 1}$$

where "." is the inner product of two vectors, $||\ ||$ is the magnitude of a vector. Cosine Similarity produces a value between -1 and 1, where 1 signifies perfect similarity, 0 implies no similarity (orthogonal vectors), and -1 indicates complete dissimilarity. In the context of our analysis, we calculated the Cosine Similarity between pairs of poem BoWs to determine how similar or dissimilar they are based on the trigrams used within the poems.

## 3.4. Topic Modeling with ParsBERT Word Embedding

Topic modeling is a BoW-based unsupervised learning method to systematically categorize, structure, and identify themes in texts. It is a statistical model to discover topics in a set of documents. Latent Dirichlet Allocation is a renowned topic modeling technique in NLP [40]. LDA assumes that documents (e.g., each book) are a combination of topics and topics are a mixture of tokens (or words). Once the preprocessing is complete including tokenization, LDA is an iterative process where it starts assuming there are *k* topics in the corpus. It first iterates through all the documents and randomly assigns each word in the document to one of the *k* topics. Then for each document, it iterates through each word *w* and calculates two probabilities: 1) the probability of topic *t* in document *d* (the proportion of words in document *d* that are assigned to topic *t*). This captures how many words belong to topic *t* for a given document *d* excluding the current word. If many words from document *d* belongs to topic *t*, it is highly likely that word *w* belongs to topic *t*. 2) the probability of word *w* in topic *t* (the proportion of assignments to topic *t* across all documents that are generated by this word *w*). This captures how many documents are in topic *t* because of word *w*. The multiplication of these probability is assigned as the probability of word *w* for topic *t*, and thus, the probability topic *t* in document *d* is updated accordingly. This is an iterative process, and it continues until convergence.

LDA generates a probabilistic vector that is commonly used in topic modeling. The size of this vector depends on the number of topics chosen for each book. This vector serves as a crucial component in our analysis, aiding in the identification and categorization of topics within the poems. For our analysis, we chose the number of topics to be 4.

While LDA has been used as a prominent topic modeling method, it has limitations. LDA primarily relies on word frequency to determine topics. This approach can be limiting when dealing with Persian poetry, as it often contains words that carry nuanced meanings in different contexts. Consequently, the semantic information conveyed by the verses in the poems may not be effectively captured using LDA alone, resulting in suboptimal topic modeling outcomes.

To address this limitation and enhance the contextual understanding of the poems, we incorporated ParsBERT [35]. ParsBERT is a large language model that converts input words into vectors of size 768, thus encoding their semantic information. We used the pre-trained ParsBERT model accessible from [41]. ParsBERT is a Persian language model based on BERT architecture.

It consists of transformer encoder layers with self-attention mechanisms, enabling it to understand complex patterns in Persian text. ParsBERT is bidirectional and trained using masked language modeling to capture context dependencies effectively. It incorporates segment embeddings for sentence-level tasks. Pretrained on a large Persian corpus, encompassing various Persian text sources, such as Persian Wikipedia[1], BigBangPage[2], Chetor[3], Eligasht[4], Digikala[5], Ted Talks subtitles[6], several fictional books and novels, and more than 250 Persian news websites, ParsBERT performs well in different tasks including text classification, NER, and sentiment analysis. As the input to the model consists of poems, we used the average of the vectors of individual words within each poem. This process allows us to create a unified vector representation specific to each poem, capturing its overall semantic essence [35].

Inspired by [42], to incorporate both analytic and semantic information provided by LDA and ParsBERT respectively, we employ an autoencoder for feature fusion. Initially, the vectors obtained from LDA (size=4) and ParsBERT (size=738) are concatenated and then fed into the autoencoder. However, before performing the concatenation, we introduce a trade-off parameter, denoted as $\alpha$, to strike a balance between the feature vectors representing analytic and semantic information. To achieve this trade-off, the LDA vectors are multiplied by $\alpha$. Consequently, the value of $\alpha$ determines the extent to which we rely on the analytic information for the topic modeling process. A higher value of $\alpha$ places greater emphasis on the analytic information extracted by LDA method. For our experiments, we set $\alpha$=15.

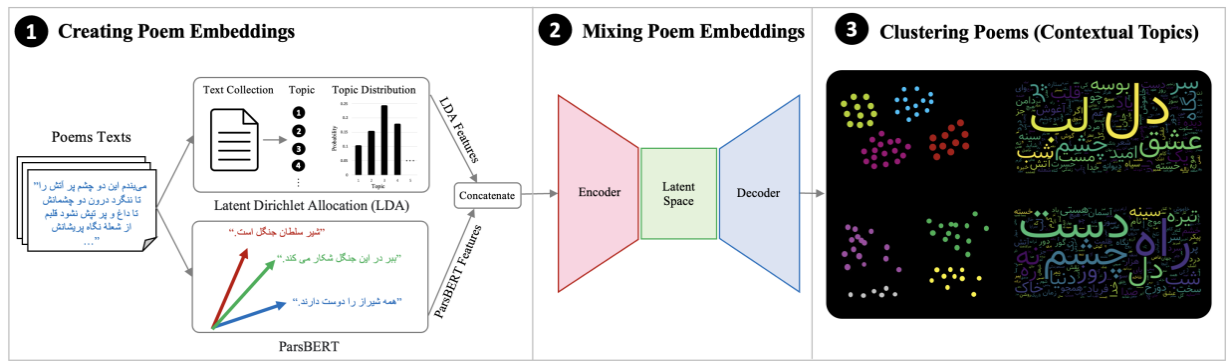

Figure 1- Overview of our proposed LDA+ParsBERT pipeline

The autoencoder is trained for 1000 epochs with a batch size of 128 using all five books from Forough's poetry collection. It is designed with a single hidden dense layer comprising 16 neurons. As a result, an input vector, originally of size 768 plus the number of topics (4), is transformed into a condensed vector of size 16. The latent space of the autoencoder is then utilized as the fused feature vector, effectively combining the two contextual information sources. The decision to train the autoencoder with all poems is because of the small sample size of poems in each individual book. It's important to note that the autoencoder is solely utilized for dimensionality reduction and mixing the embeddings from LDA and ParseBert models. Thus, the training can be done exclusive to any specific book in the dataset.

---

[1] https://dumps.wikimedia.org/fawiki/
[2] https://bigbangpage.com/
[3] https://www.chetor.com/
[4] https://www.eligasht.com/Blog/
[5] https://www.digikala.com/mag/
[6] https://www.ted.com/talks

By leveraging the autoencoder, we can integrate both analytic and semantic aspects of the texts, resulting in a more comprehensive representation of the data for topic modeling purposes. Figure 1 illustrated our proposed pipeline.

Once the fused feature vectors were created for each poem, K-means clustering algorithm was applied to the latent space representations of poems in each book, followed by PCA analysis for visualization. Finally, the word cloud for the clusters is represented as the topic for each book.

## 4. Results and Analysis

### 4.1. Frequency Analysis

Figure 2 show the results of the frequency analysis for each book.

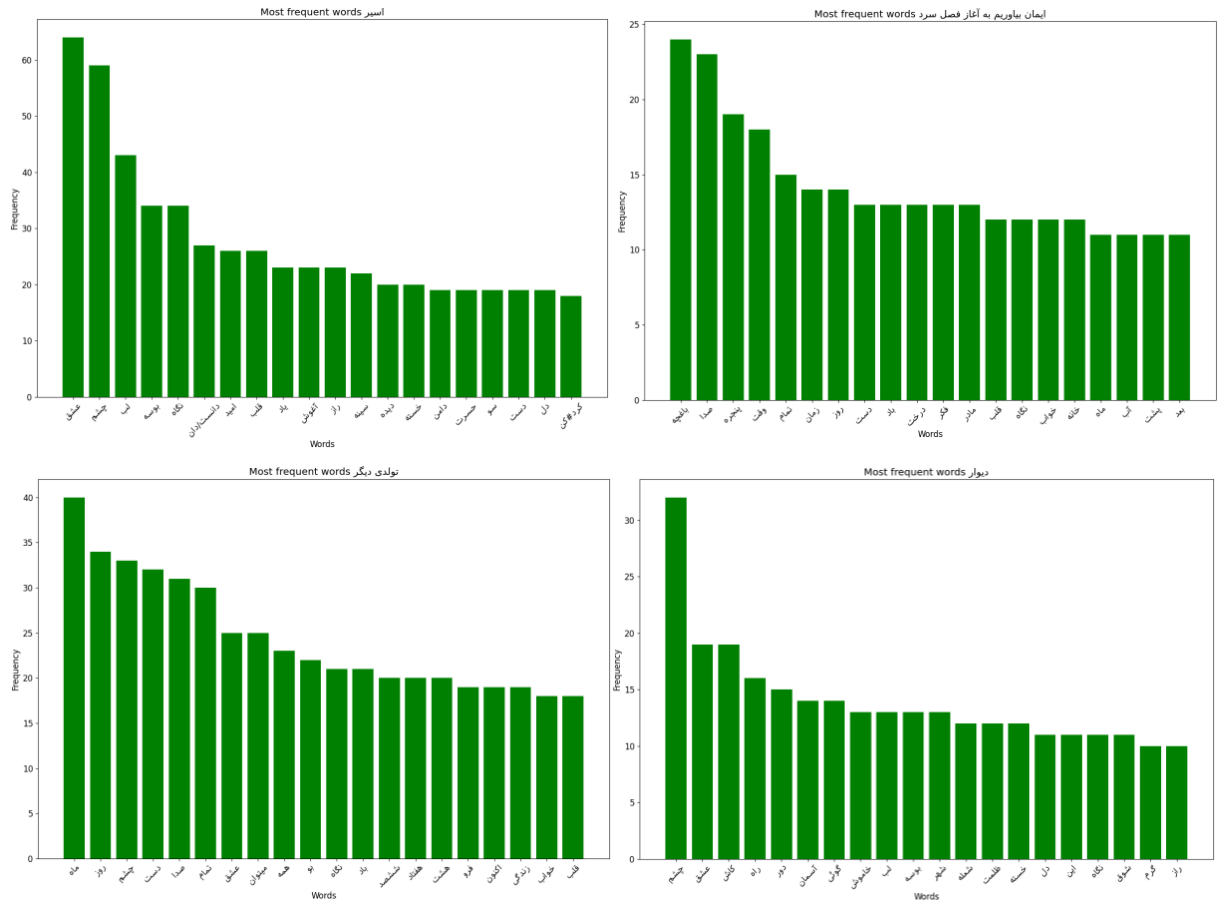

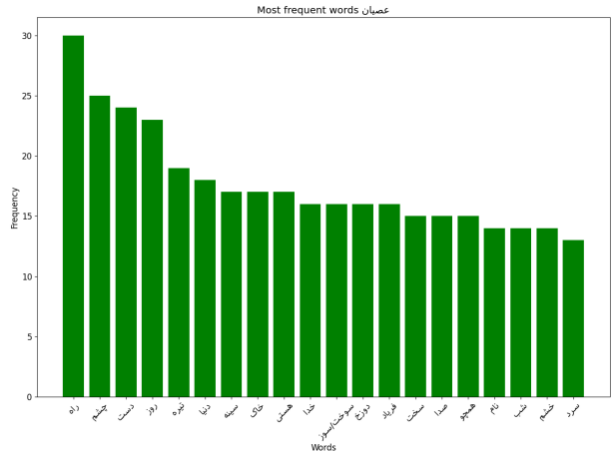

Figure 2- Word frequency analysis across five works of Forough Farrokhzad: a comparative histogram visualization

Figure 3 illustrates word clouds where the size of each word represents the frequency of the word in the book.

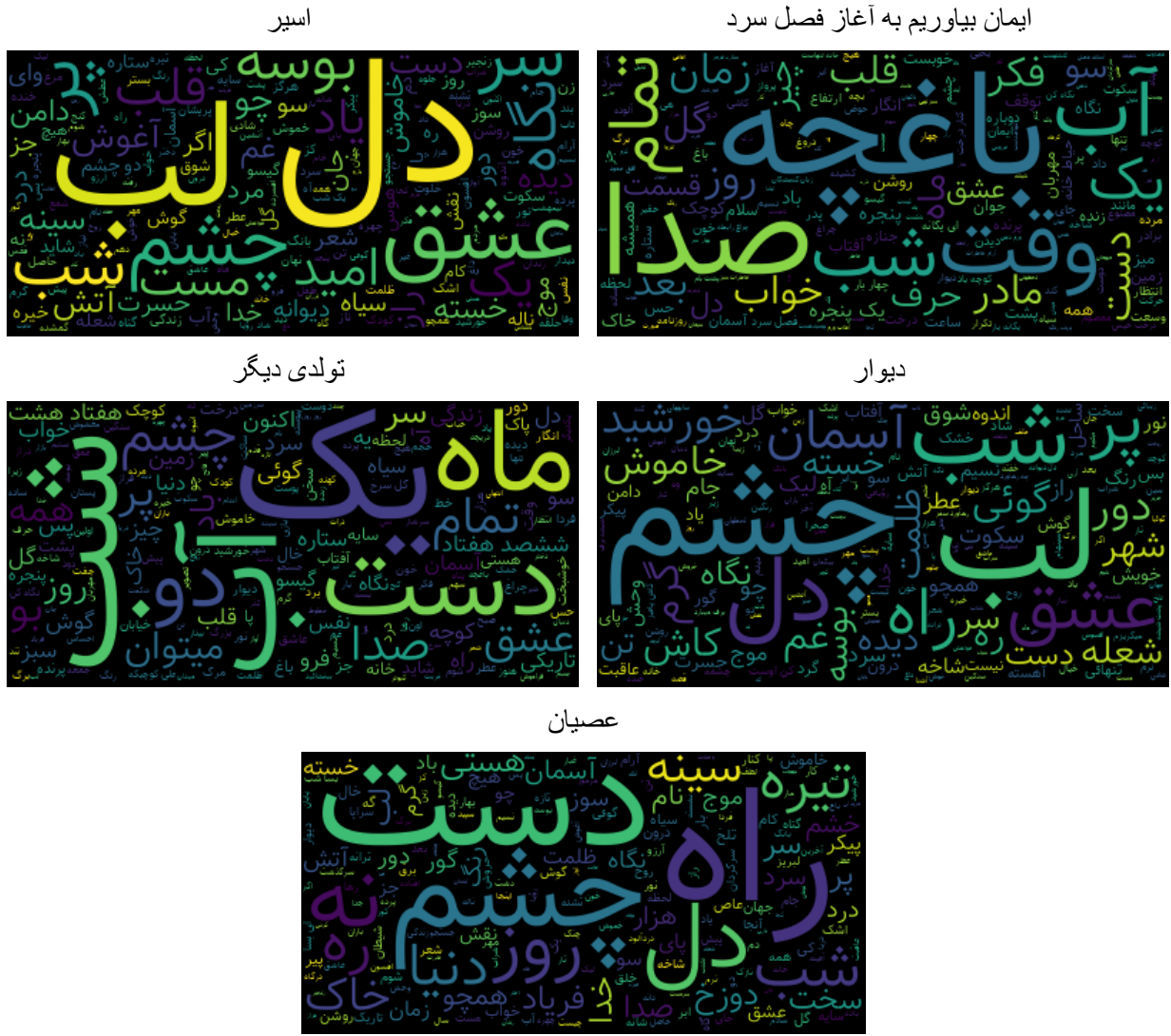

Figure 3- Word Clouds of Five Pivotal Works by Forough Farrokhzad

## 4.2. Clustering

### 4.2.1. Frequency-Based Clustering (Top five most common words)

Figure 4 show the results of the Frequency Clustering, using the top five most common words in each poem. For each book, the plot illustrates the PCA results using two principal components. The table shows the clustering results using the entire data (no dimensionality reduction).

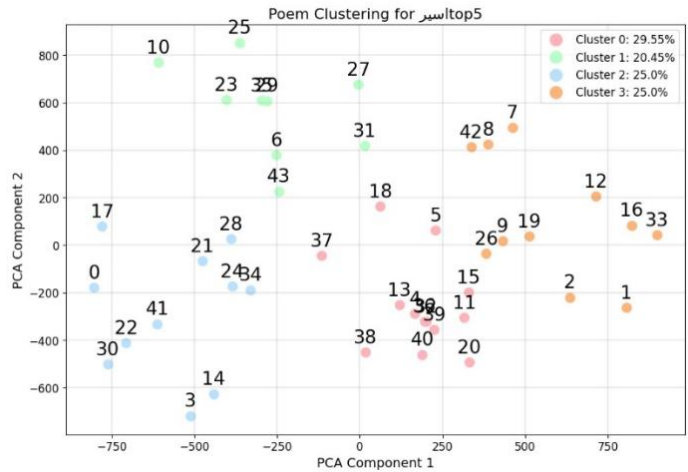

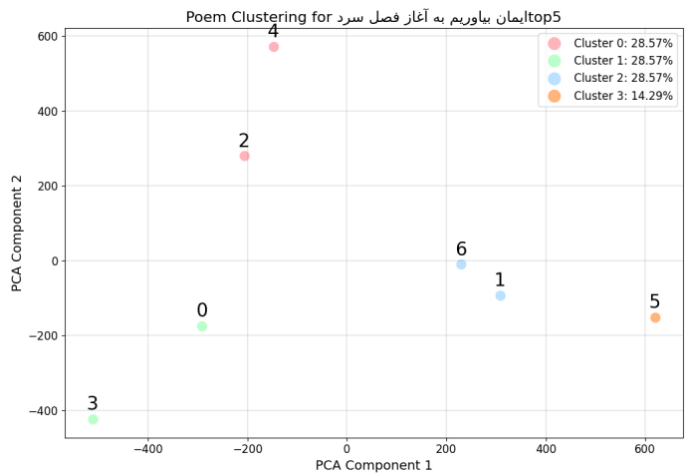

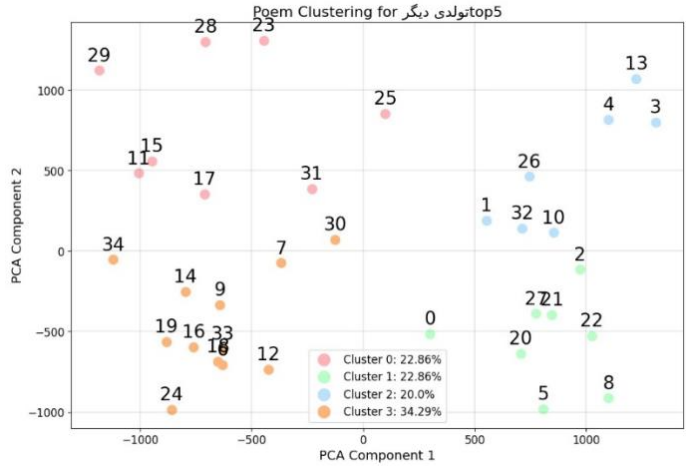

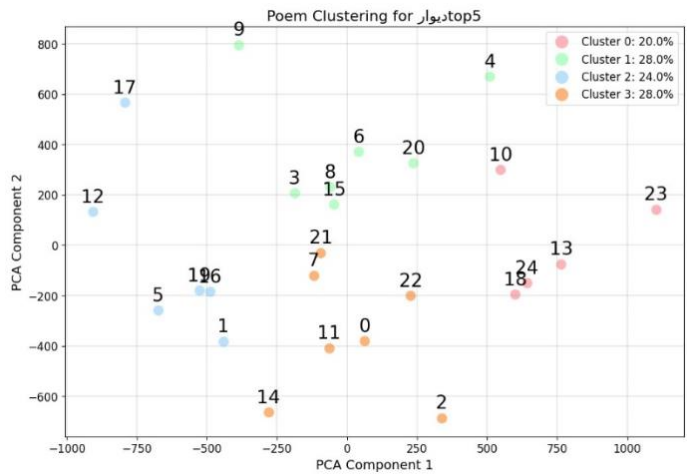

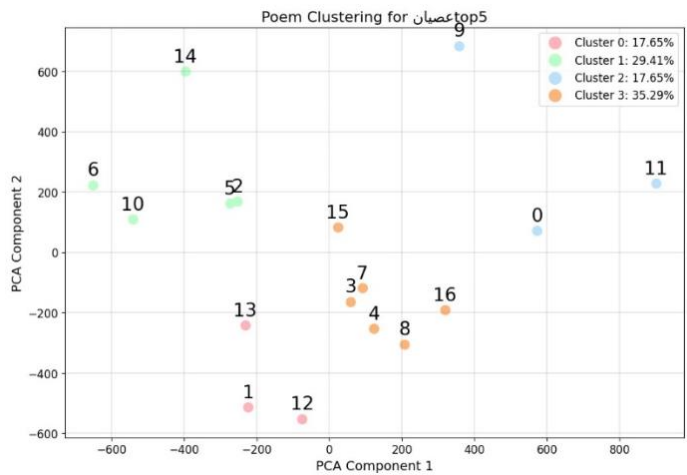

Figure 4- Clustering results using the top five most common words in each poem

## 4.2.2. N-Gram Frequency-Based Clustering

Figure 5 show the results of the Frequency Clustering, using the trigrams in each poem. For each book, the plot illustrates the PCA results using two principal components. The table shows the clustering results using the entire data (no dimensionality reduction).

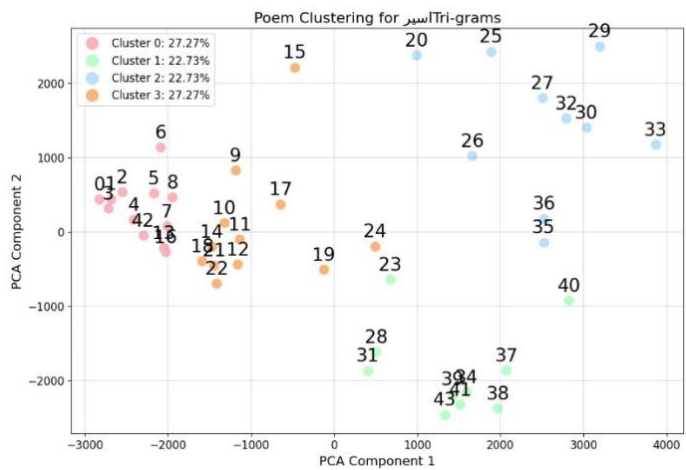

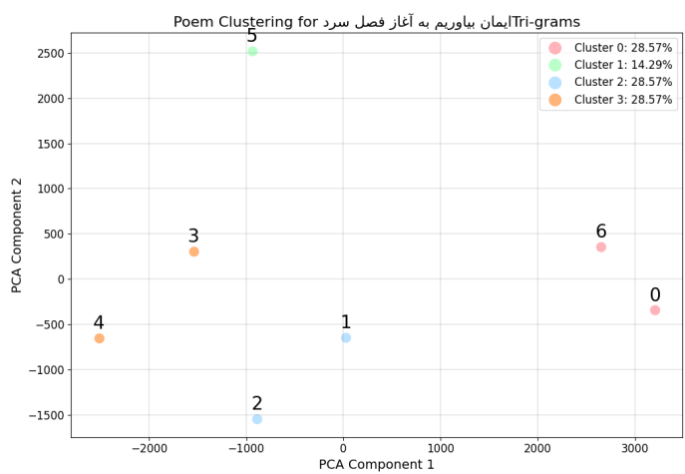

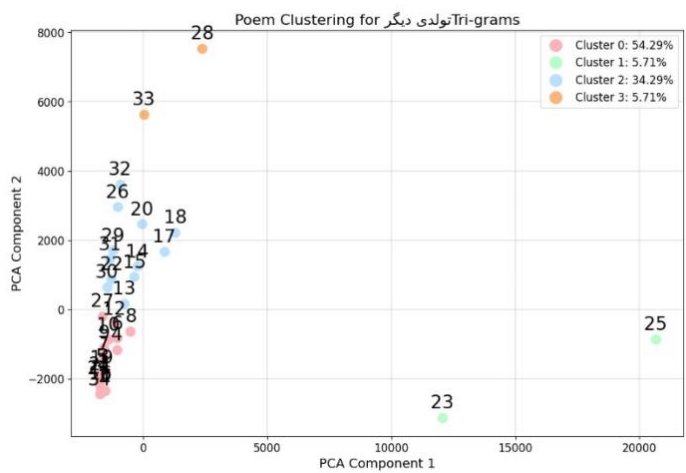

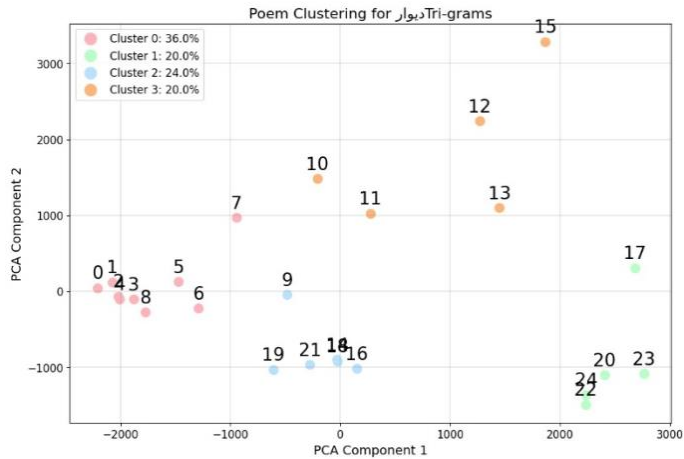

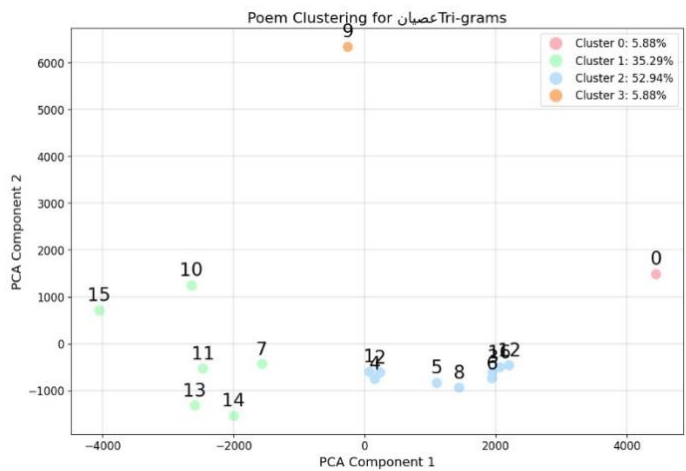

Figure 5- Clustering results using the trigrams in each poem

### 4.2.3. Similarity Analysis

Figure 6 shows the similarity matrix of each book based on the Cosine metric, calculated using the embeddings from the trigrams of the poems.

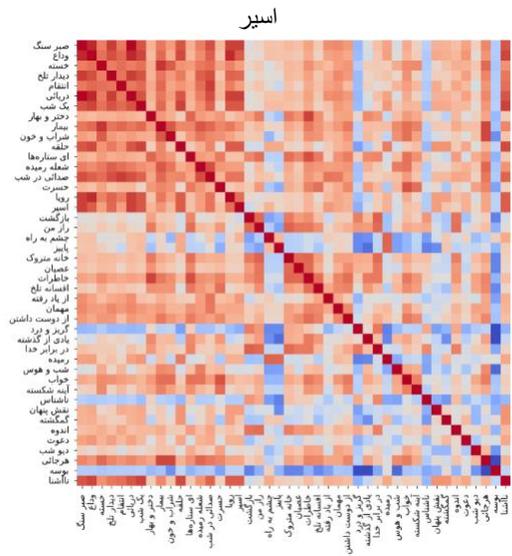
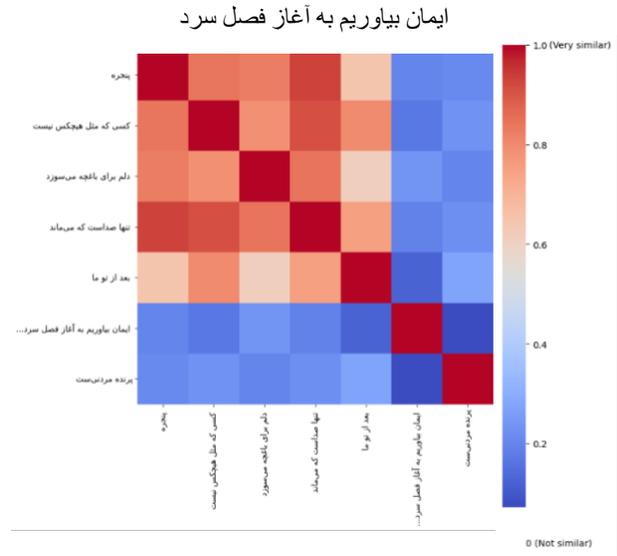
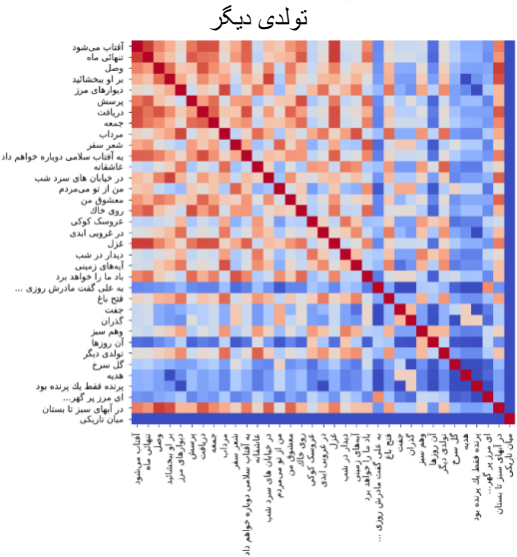
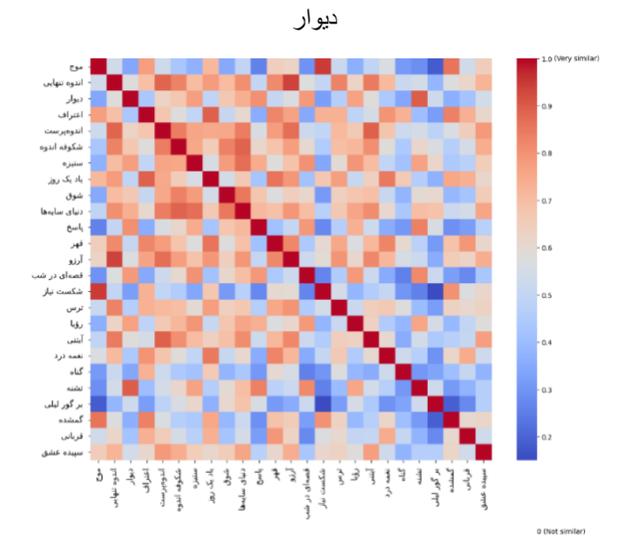
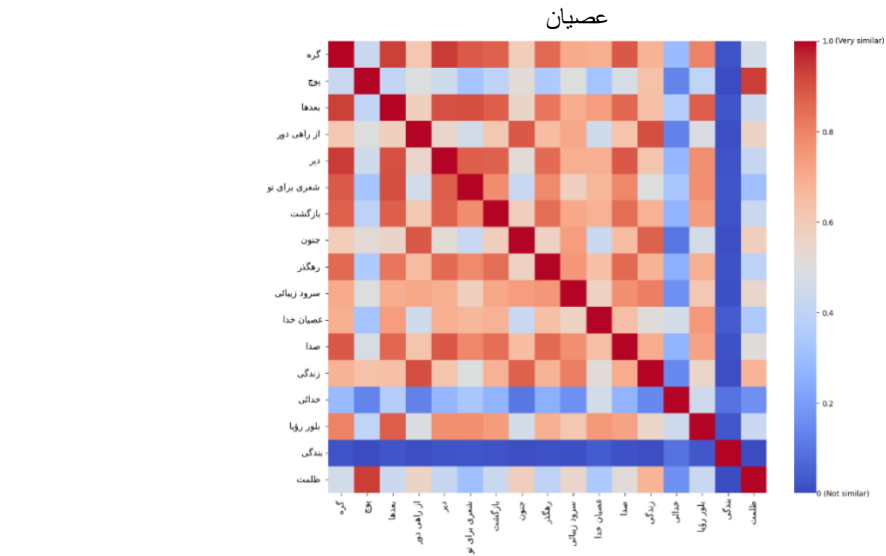

Figure 6- Heatmap of Cosine Similarity Scores the of Poems in the five studied books

## 4.3. Results of Topic Modeling with ParsBERT Word Embedding

In this section, the results of topic modeling with ParsBERT word embedding are illustrated. A visualization of the clusters within each book is depicted in Figure 7. We employed the Uniform Manifold Approximation and Projection (UMAP) [43] to reduce the dimensionality of the embeddings of each poem to 2 dimensions for illustrative purposes. UMAP is a powerful technique that effectively captures the underlying structure of high-dimensional data [43]. In Figure 7, it is evident that some books, such as "اسیر", exhibit four distinct clusters in their embeddings. However, for certain other books, the number of clusters (or topics) appears to be a somewhat arbitrary estimation. It is essential to recognize that while UMAP excels in preserving distances following dimensionality reduction, the distances between poem embeddings in low dimensions may not precisely reflect their actual distances in the original high-dimensional space. Therefore, the representation shown in the corresponding tables utilizes the full latent feature vector, which is more reliable.

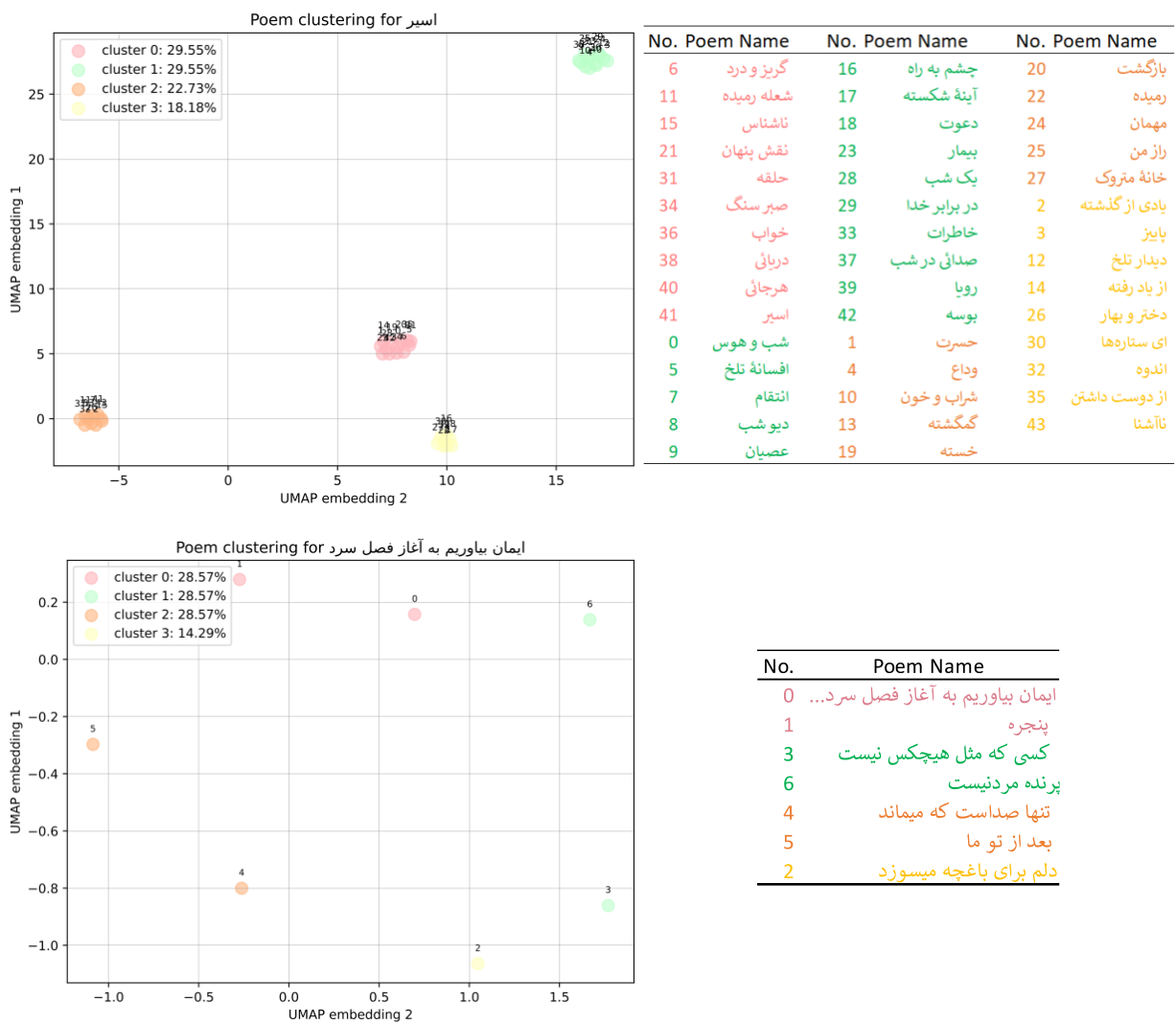

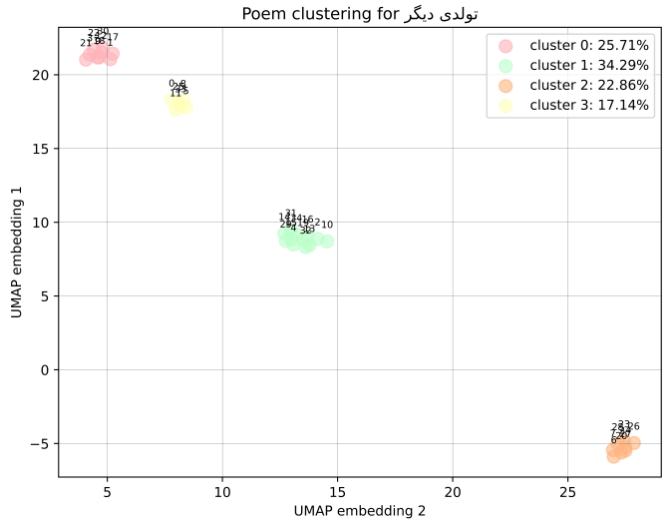

| No. | Poem Name | No. | Poem Name |
|---|---|---|---|
| 1 | بر او ببخشائید | 31 | باد ما را خواهد برد |
| 3 | وصل | 32 | غزل |
| 9 | تنهایی‌ماه | 34 | میان تاریکی |
| 12 | در خیابان های سرد شب | 6 | دیوارهای مرز |
| 17 | دیدار در شب | 7 | جمعه |
| 18 | وهم سبز | 20 | فتح باغ |
| 21 | گل سرخ | 23 | به علی گفت مادرش روزی... |
| 22 | آفتاب میشود | 24 | پرنده فقط یک پرنده بود |
| 30 | شعر سفر | 26 | به آفتاب سلامی دوباره خواهم داد |
| 2 | دریافت | 27 | من از تو میمردم |
| 4 | عاشقانه | 28 | تولدی دیگر |
| 10 | معشوق من | 0 | آن روزها |
| 13 | در غروبی ابدی | 5 | پرسش |
| 14 | مرداب | 8 | عروسک کوکی |
| 15 | آیه‌های زمینی | 11 | گذران |
| 16 | هدیه | 25 | ای مرز پر گهر... |
| 19 | جفت | 33 | در آبهای سبز تابستان |
| 29 | روی خاک | | |

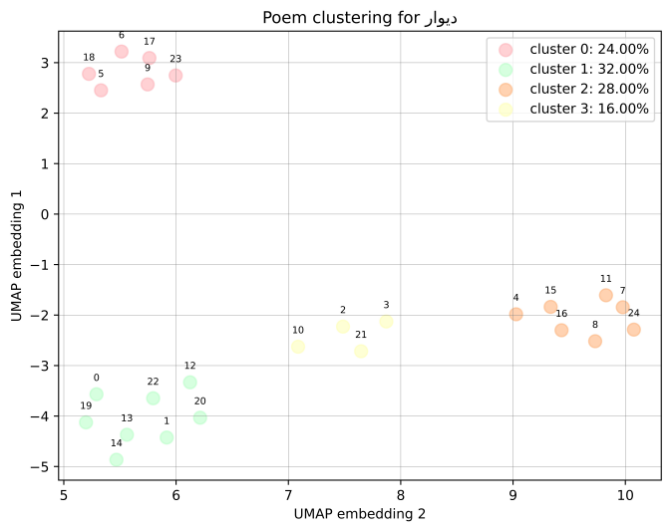

| No. | Poem Name | No. | Poem Name |
|---|---|---|---|
| 5 | شوق | 22 | آرزو |
| 6 | اندوه تنهایی | 4 | موج |
| 9 | شکوفهٔ اندوه | 7 | قصه ای در شب |
| 17 | دنیای سایه ها | 8 | شکست نیاز |
| 18 | نغمهٔ درد | 11 | رؤیا |
| 23 | آبتنی | 15 | تشنه |
| 0 | گناه | 16 | ترس |
| 1 | بر گور لیلی | 24 | سپیدهٔ عشق |
| 12 | دیوار | 2 | اعتراف |
| 13 | ستیزه | 3 | یاد یک روز |
| 14 | قهر | 10 | پاسخ |
| 19 | گمشده | 21 | قربانی |
| 20 | اندوه پرست | | |

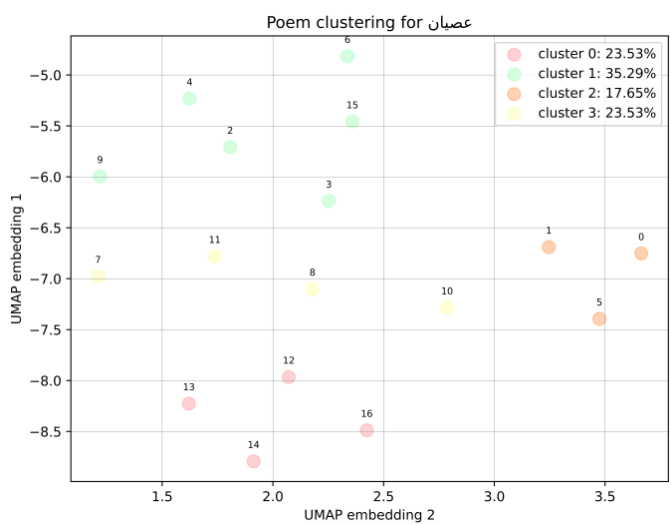

| No. | Poem Name | No. | Poem Name |
|---|---|---|---|
| 12 | پوچ | 15 | بلور رؤیا |
| 13 | دیر | 0 | بندگی |
| 14 | صدا | 1 | گره |
| 16 | ظلمت | 5 | سرود زیبایی |
| 2 | بازگشت | 7 | بعدها |
| 3 | از راهی دور | 8 | زندگی |
| 4 | رهگذر | 10 | عصیان خدا |
| 6 | جنون | 11 | شعری برای تو |
| 9 | خدایی | | |

Figure 7- The result of clustering the books for topic modeling based on ParsBERT and LDA. The UMAP technique is used for the dimensionality reduction and illustration in 2 dimensions.

## 4.4. Word Cloud Results for Topic Modeling with ParsBERT

In this section we present the results of the proposed topic modeling with a predetermined topic number of 4. For content analysis, we created word cloud visualization models for each book, i.e., a word cloud for each topic in a specific book. Word cloud visualization will give an idea of what words are often used in specific text. Figure 8 to Figure 12Figure *8* show the word clouds for each topic in each book.

Topic 0          Topic 1

Topic 2          Topic 3

Figure 8- Word Clouds of اسیر

| Topic 0 | Topic 1 |
| --- | --- |
| Topic 2 | Topic 3 |

Figure 9- Word Clouds of ایمان بیاوریم به آغاز فصل سرد

| Topic 0 | Topic 1 |
| --- | --- |
| Topic 2 | Topic 3 |

Figure 10- Word Clouds of تولدی دیگر

| Topic 0 | Topic 1 |
|---|---|
| 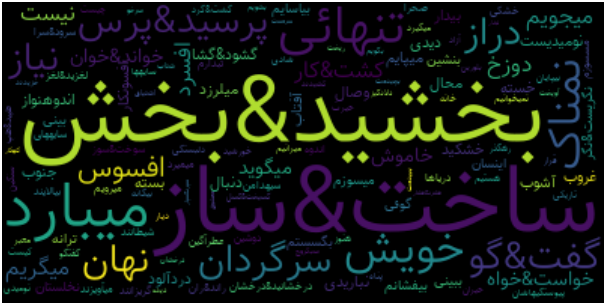 | 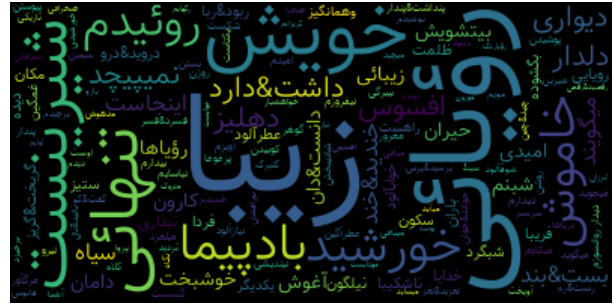 |
| Topic 2 | Topic 3 |
| 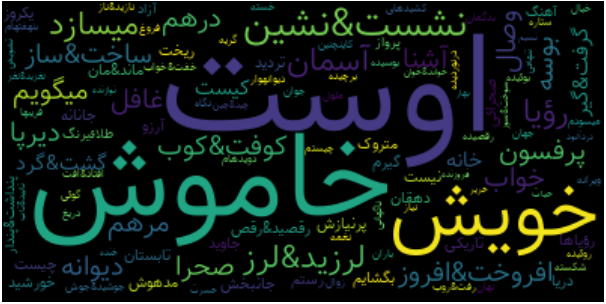 | 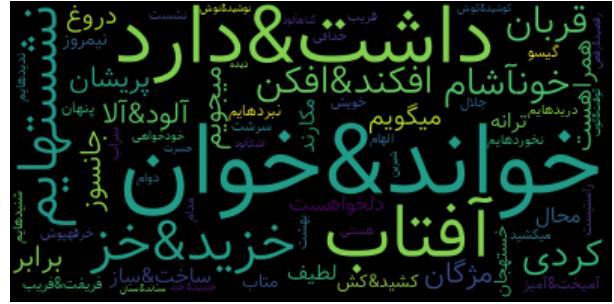 |

Figure 11- Word Clouds of دیوار

| Topic 0 | Topic 1 |
|---|---|
| 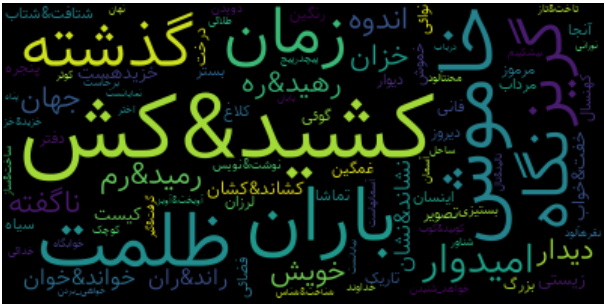 | 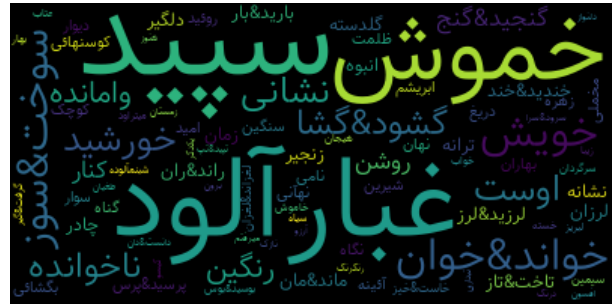 |
| Topic 2 | Topic 3 |
| 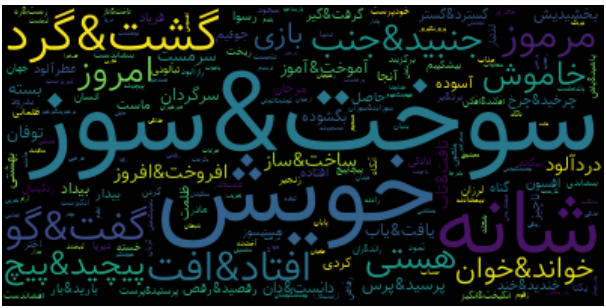 | 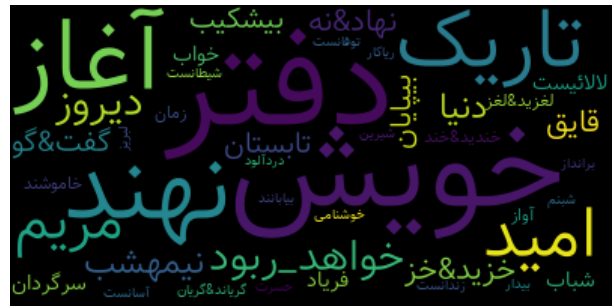 |

Figure 12- Word Clouds of عصیان

# 5. Discussion & Conclusion

Our study was conducted to deeply explore and analyze the works of Forough Farrokhzad, a seminal female figure in modern Persian poetry, through the lens of NLP and AI. By employing advanced NLP and AI techniques, our goal was to explore the potential of NLP for uncovering thematic, stylistic, and linguistic patterns within her poetry. Farrokhzad's work, known for its emotive depth and distinctive style, presents a fertile ground for such an analysis, allowing us to explore the interplay of language, emotion, and cultural context in Persian poetry.

Our methodology involved a comprehensive analysis of five books by Farrokhzad, using a combination of frequency analysis, clustering, including the top five most frequent words and trigram analysis, and topic modeling with LDA and ParsBERT. The initial steps included preprocessing the text to remove stop words and normalizing the words for analysis. We utilized frequency analysis to identify the most common words and trigrams, establishing a foundational understanding of recurring themes through clustering of the poems in each book. Acknowledging the limitations of this approach in capturing the poems' deeper layers, we transitioned to more advanced techniques, including trigram analysis, to probe into more complex patterns and offer a richer understanding of Farrokhzad's poetic artistry. The subsequent multifaceted application of LDA and ParsBERT methods allowed for a deeper semantic exploration through a more coherent and nuanced clustering, which revealed patterns and similarities among poems with deeper semantic meaning and less overlap, indicating a higher degree of similarity within each cluster.

This approach, while foundational, set the stage for deeper investigations using advanced methods like LDA and BERT, which unraveled more complex semantic patterns. The interplay between these methodologies enriched our analysis, offering a comprehensive view of Farrokhzad's work. It also highlighted the challenges in evaluating unsupervised learning techniques in literature. The analysis of digital humanities libraries, particularly in the context of Forough's books, presents many challenges underscored by the inherent complexities of applying machine learning techniques. The absence of definitive ground-truth data compounds these challenges, particularly in domains where subjectivity and interpretation are paramount. Consequently, establishing a concrete benchmark for analysis becomes notably arduous, necessitating a reliance on unsupervised learning approaches.

In our methodology, algorithms autonomously identify patterns and structures within the text corpus, granting us flexibility and adaptability in analysis. However, this approach also highlights the imperative need for careful validation and interpretation of results, especially as we navigate the delicate balance between computational insights and the nuanced intricacies of literary interpretation.

In our exploration of clustering techniques and the generation of word clouds for each book, we observed distinct differences in the effectiveness of various methodologies. ParsBert plus LDA method, for instance, emerged as a more robust approach, producing clusters of poems imbued with deeper semantic meaning compared to the frequency analysis method. This enhanced capability to consider the semantics of the poems led to more meaningful inter-cluster relationships. For instance, Fig. 7 demonstrate how clustering using ParsBert and LDA resulted in less overlap between clusters for majority of the books suggesting a higher degree of similarity within each cluster compared to those produced by trigram or frequency analysis (Figures 4 and 5).

The findings of this study have several implications. First, they demonstrate the potential of AI and NLP in literary studies, especially for languages and literature that are underrepresented in digital humanities research. The improved clustering and topic modeling techniques can offer new insights into literary texts, revealing underexplored themes and stylistic features. Furthermore, the ability to quantify similarities among poems provides a novel approach to studying literary corpora, potentially leading to a better understanding of an author's thematic preoccupations and stylistic evolution.

This work has limitations. In our analysis of Forough Farrokhzad's poetry using unsupervised learning, evaluating the methods is challenging due to the absence of predefined outcomes or hypotheses. Unsupervised learning explores data patterns without explicit instructions, making it difficult to assess accuracy or effectiveness against a specific standard. This is particularly complex in literary studies, where interpretations are subjective and multifaceted. In addition, some hyperparameters of the unsupervised learning algorithms are set heuristically, such as the number of clusters. A different number of clusters (3 or 5) may lead to different clustering results. Thus, unsupervised learning methods are sensitive to such hyperparameters, which may make the results less reliable. Finally, this work presents an exploration of Persian poetry with NLP with no clear hypothesis in mind making the evaluation of the results a challenging task.

While our current work has successfully identified topics within each book through poem clustering, it may lack the granularity necessary to capture finer connections. Hence, our future endeavors will focus on refining our approach to uncover topics within each book in a more fine-grained manner. By delving deeper into the intricate web of themes and motifs present in Forough Farrokhzad's poetry, we aim to enrich our understanding of her literary legacy within the digital realm and uncover hidden nuances that traditional methodologies may overlook. Additionally, establishing a more concrete ground truth will be crucial for validating our computational analysis of Forough Farrokhzad's poetry. Collaborating with literary experts and scholars to annotate themes, motifs, and stylistic elements within a curated subset of her poems could provide this essential benchmark.